\pdfoutput=1
\documentclass[11pt]{article}

\usepackage[final]{EMNLP2022}

\usepackage{times}
\usepackage{latexsym}
\usepackage{subcaption}
\usepackage{amsmath,amsfonts,amssymb}
\usepackage{breqn}
\usepackage{tabularx}
\usepackage{multirow}
\usepackage{graphicx}
\usepackage{color}
\usepackage{tcolorbox}
\usepackage{CJK}
\usepackage{adjustbox}
\usepackage{xcolor}
\usepackage{colortbl}
\usepackage{multicol}
\usepackage{vwcol} 
\newsavebox{\algleft}
\newsavebox{\algright}
\usepackage{bm}
\usepackage{booktabs}
\usepackage{ctable} % for \specialrule command
\usepackage{amsmath,stackengine}
\usepackage[linesnumbered,ruled,vlined]{algorithm2e}
\usepackage{times}
\usepackage{latexsym}
\usepackage{bbm}
\usepackage[utf8]{inputenc} % allow utf-8 input
\usepackage[T1]{fontenc}    % use 8-bit T1 fonts
\usepackage{hyperref}       % hyperlinks
\usepackage{url}            % simple URL typesetting
\usepackage{booktabs}       % professional-quality tables
\usepackage{amsfonts}       % blackboard math symbols
\usepackage{nicefrac}       % compact symbols for 1/2, etc.
\usepackage{microtype}      % microtypography
\usepackage{xcolor}         % colors
\usepackage{bm}
\usepackage{microtype}

\usepackage[T1]{fontenc}
\usepackage[utf8]{inputenc}

\usepackage{microtype}

\usepackage{inconsolata}

\title{Continual Training of Language Models for Few-Shot Learning}
\author{
Zixuan Ke$^{1}$, Haowei Lin$^{2}$, Yijia Shao$^{2}$, Hu Xu$^{1}$, Lei Shu$^{1}$\thanks{~~Now at Google Research  \url{leishu@google.com}}~~~and Bing Liu$^{1}$\\ 
$^1$Department of Computer Science, University of Illinois at Chicago\\
$^2$Wangxuan Institute of Computer Technology, Peking University\\
$^1$\texttt{\{zke4,hxu48,liub\}@uic.edu}\\  $^2$\texttt{\{linhaowei, shaoyj\}@pku.edu.cn} \\
}

\begin{document}
% Zixuan Ke, Haowei Lin, Yijia Shao, Hu Xu, Lei Shu and Bing Liu 

\maketitle
\begin{abstract}
{\color{black}%This paper studies continual pretraining a language model (LM) to accumulate domain specific knowledge. It has been shown that fine-tuning a pre-trained LM is one of the most effective approaches to learn an end-task.
% \hu{the first sent. is harder for me to parse, probably because 'adapt' and 'training' is unclear what it refers to, how about 'It is well-known that training (a.k.a., post-training/pre-finetuning) a language model (LM) separately on multiple unlabeled domain corpus in-between pre-training and fine-tuning can achieve significantly better results than direct fine-tuning.'} 
{\color{black}Recent work on applying large language models (LMs) achieves impressive performance in many NLP applications.~Adapting or post-training an LM using an unlabeled domain %\zixuan{reviews: What does unlabeled corpora mean in the context of language models?} 
corpus can produce even better performance for end-tasks in the domain. % them with However, these models still perform poorly on a wide range of domains outside of their respective training and evaluation sets. To address this limitation, 
This paper proposes the problem of % extending its knowledge and abilities, without forgetting previous skills. In this paper, we propose a 
continually extending an LM by incrementally post-train the LM with a sequence of unlabeled domain corpora to expand its knowledge without forgetting its previous skills. The goal is to improve the few-shot end-task learning in these domains.  
The resulting system is called CPT (\textit{C}ontinual \textit{P}ost-\textit{T}raining), which to our knowledge, is the first continual post-training system. Experimental results verify its effectiveness.\footnote{{\color{red}\url{https://github.com/UIC-Liu-Lab/CPT}}}}
%and evaluate the performance by the corresponding end-tasks. Our resulting model, CPT, can continually post-trains on different domains \textit{with no forgetting} of previous skills and improve the \textit{few-shot} learning performance in the learned domains. 
 %system of the proposed paradigm. 

% It is well-known that if we \textit{post-train} (a.k.a., \textit{adapting} or \textit{pre-finetuning}) a language model (LM) with an unlabeled domain corpus before end-task fine-tuning in the domain, we can achieve significantly better results than without post-training. This paper proposes a new paradigm that continually post-trains an LM with a sequence of unlabeled domains. This is crucial because new domain data keep emerging and the LM should be able to continually learn new information. The proposed approach, CPT, can continually post-trains on different domains without forgetting the previous ones and improve the few-shot performance in the learned domains. To our knowledge, CPT is the first system in the proposed paradigm.  %} The resulting adapted LM can be used by any end-tasks in these domains through fine-tuning. 
% \zixuan{The sentence may be too long.  "Complementing the effort of building ever powerful LMs, we propose a new paradigm that continually adapts a general LM with corpora of many domains. This can help accumulate domain-specific knowledge. After continually domain-adaptive pre-training, any end-tasks in these domains can make use of the pre-trained LM through fine-tuning." }\footnote{The code has been uploaded as part of Appendix.} 
}

\end{abstract}

%\zixuan{

%main paper: neurips_2021.tex; 

%Supplentary: neurips_2021_Appendix.tex;

% Previous author response: icml2021_author_response.tex}

\section{Introduction}
\label{sec.intro}

%Pre-trained language models have been instrumental to the major advance of NLP for the past few yearsr~\cite{X}. It has prompted the community to build ever larger and more powerful language models.
%\ls{would it be better to mention post training before finetuning? finetuning is not our goal, it is just one way for evaluation. like: post training has been widely studied to reduce the domain-/task- discrepancy between pre-trained language model and end-task.}
%Fine-tuning a pretrained language model (LM) has been shown to be one of the most effective approaches to learning an end-task~\cite{DBLP:conf/naacl/XuLSY19,sun-etal-2019-utilizing}. 

% This work proposes a continual post-training (CPT) technique to continually train a single pretrained language model using corpora in many domains, and subsequently finetune for an end-task specific to a domain. CPT works by training adapter-like modules that are inserted in the transformer layers of RoBERTa and solely trained using the domain data. To retain domain information, domain-specific masks are applied to the adapter modules and learned end-to-end. The results show that the proposed approach outperforms other baselines and a naive continual learning baseline on four text classification tasks in the few-shot setting.

Recent work has shown that large LMs have the ability to perform few-shot (or even zero-shot) learning well \cite{DBLP:conf/nips/BrownMRSKDNSSAA20,DBLP:journals/corr/abs-2112-11446,DBLP:journals/corr/abs-2201-11990}. 
{\color{black}\textit{Post-training} (a.k.a., \textit{domain-adaptive pre-training} or \textit{pre-finetuning}) an LM with a large unlabeled domain corpus before end-task fine-tuning in the domain achieves better results~\cite{DBLP:conf/naacl/XuLSY19,gururangan2020don} than directly fine-tuning the  LM.
% However, they are weak outside of \bing{how to know `outside'}\zixuan{we don't know (also not the key in this paper). The only thing we can do is just show the without post-train is worse.}their training and evaluation datasets (as shown in Sec.~\ref{sec:results}). 
This paper goes a step further to study the problem of improving an LM's ability to handle new and ever emerging domains. For this, one needs to \textit{continually post-train} the LM with a sequence of domains. A key issue associated with this problem is \textit{catastrophic forgetting} (CF).\footnote{
CF means that learning a new task/domain may need to modify the existing network, which degrades the performance of previous tasks/domains~\cite{mccloskey1989catastrophic}.} This paper thus investigates how to continually extend the LM's knowledge without suffering from CF. {\color{black}From a broader perspective, since training a large LM from scratch is extremely expensive and computation intensive, incrementally updating the LM with the latest language data reflecting the ever changing development of the language itself, social events and the knowledge from different fields is becoming more and more critical. As humans are very effective at incremental learning, if we can imitate this human capability with little or no forgetting, we will be pushing the AI research forward significantly.}   

  % Note that post-training is called \textit{pre-finetuning} and \textit{domain-adaptive pretraining} in the literature. 
The proposed system, called CPT, is a  %\textit{Continual Post-\textit{T}raining} (CPT) (
continual learning (CL) system for post-training. Starting from a pre-trained LM (e.g., RoBERTa \cite{DBLP:journals/corr/abs-1907-11692}), it incrementally post-trains the LM with a sequence of domains using their unlabeled corpora. Once a task (a domain in our case) \footnote{We will use the term \textit{domain} in this paper to be consistent with the post-training literature} is trained, its data is no longer accessible. At any time, the resulting continually post-trained LM can be used {\color{black}by end-tasks in the trained domains.} % \zixuan{claim "any end-task" is risky, because we only test on one few-shoe end task} 
This is in the \textit{task-incremental learning} (TIL) setting of CL, where the task id (domain id in our case) is provided when the learned model of a task needs to be used later (the use of domain id is discussed in Sec.~\ref{sec:task_mask}).\footnote{CL has two other settings: \textit{class-incremental learning} and \textit{domain-incremental learning}~\cite{Ven2019Three}.}  This paper proposes an effective approach called CPT and focuses on the  challenging and practical scenario of \textit{few-shot} end-task learning after post-training a sequence of domains.} 

% This paradigm is important because in the dynamic world, the data distribution shifts and new domains emerge \cite{gururangan2020don} constantly. For example, one may extend LMs pretrained over a general domain (e.g., News) to some science domains (e.g., ACL papers and AI papers\cite{DBLP:conf/acl/LoWNKW20}).\bing{We need to clearly state the difference of our work and the TsinghuaU's continual pre-trained work} This requires LMs to continually learn the new domains to achieve good performance for downstream tasks.

% This paper proposes the system CPT (\textit{Continual Post-Training}) for continual learning of post-training. CPT incrementally post-trains a single LM a sequence of domains (tasks) $d_1, d_2, ...$ using their unlabeled corpora. We use p-LM to denote the resulting LM. At any time, p-LM can be used by any end-task of any post-trained domain via its domain id. 
% The major challenge in CL is to overcome \textit{catastrophic forgetting} (CF)

%For this purpose, we first investigate the existing CL methods for conventional CL. We find that the 
Continual post-training is different from conventional CL~\cite{chen2018lifelong}. The key difference is that in conventional CL, each task is an end-task, but in our case the end-task involves fine-tuning the continual post-trained LM (called p-LM). This causes major forgetting, which we call the \textit{catastrophic butterfly effect (CBE)} and does not happen in conventional CL. Our proposed system, CPT, can solve both CF and CBE, based on a novel hard masking mechanism (Sec. \ref{sec:butterfly}) and can achieve \textit{no} forgetting. As shown in Sec. \ref{sec:results}, naively applied existing CL systems cannot effectively prevent CF (even though some existing techniques have shown almost perfect CF prevention ability in conventional CL).

Experiments in 4 domains and their corresponding end-tasks demonstrate the effectiveness of the proposed CPT system. 

\vspace{+2mm}
\noindent
{\color{black}\textbf{Related Work.} Overcoming CF is a major goal of CL~\cite{chen2018lifelong}. There are many existing approaches, e.g., 
\textit{regularization-based approaches}~\cite{Kirkpatrick2017overcoming,Seff2017continual}, 
\textit{replay-based approaches}~\cite{Rebuffi2017,Lopez2017gradient} and  \textit{parameter isolation based approaches} \cite{Serra2018overcoming,fernando2017pathnet}. {\color{black}Our CPT is based on parameter isolation and uses masks in continual post-training.} Recently, CL has drawn attention in NLP. It has been used for slot filling~\cite{shen-etal-2019-progressive}, language learning~\cite{li2019compositional}, sentence embedding~\cite{liu2019continual}, translation~\cite{khayrallah2018regularized}, cross-lingual modeling~\cite{liu2020exploring}, question answering~\cite{greco2019psycholinguistics} and text classification~\cite{DBLP:journals/corr/abs-2112-02706,ke2021Classic,sun2020lamol,huang2021continual,chuang2020lifelong,mehta2021empirical,madotto2020continual}. However, none of them tries to improve an LM.

CPT is closely related to ELLE~\cite{DBLP:conf/acl/QinZLL0SZ22}, which does \textit{continual pre-training}. The key difference is that ELLE starts from random initialization, while our CPT starts from a pre-trained LM. We tried to adapt ELLE for continual post-training by learning from a pre-trained RoBERTa but it fails to converge. This also indicates it is non-trivial to do well in our setting. Readers can refer to Appendix \ref{Sectionrelated.work} for a full coverage of the related work.
}

\section{Proposed CPT System}
\label{sec:cpt}

% \zixuan{If there is somewhere we emphasize the knowledge transfer in the model section, we need to change. I haven't checked yet.}
% \zixuan{I am thinking to add the "butterfly effect" and our simple hard masking approach}

\begin{figure}[t!]
\centering
\includegraphics[width=\columnwidth]{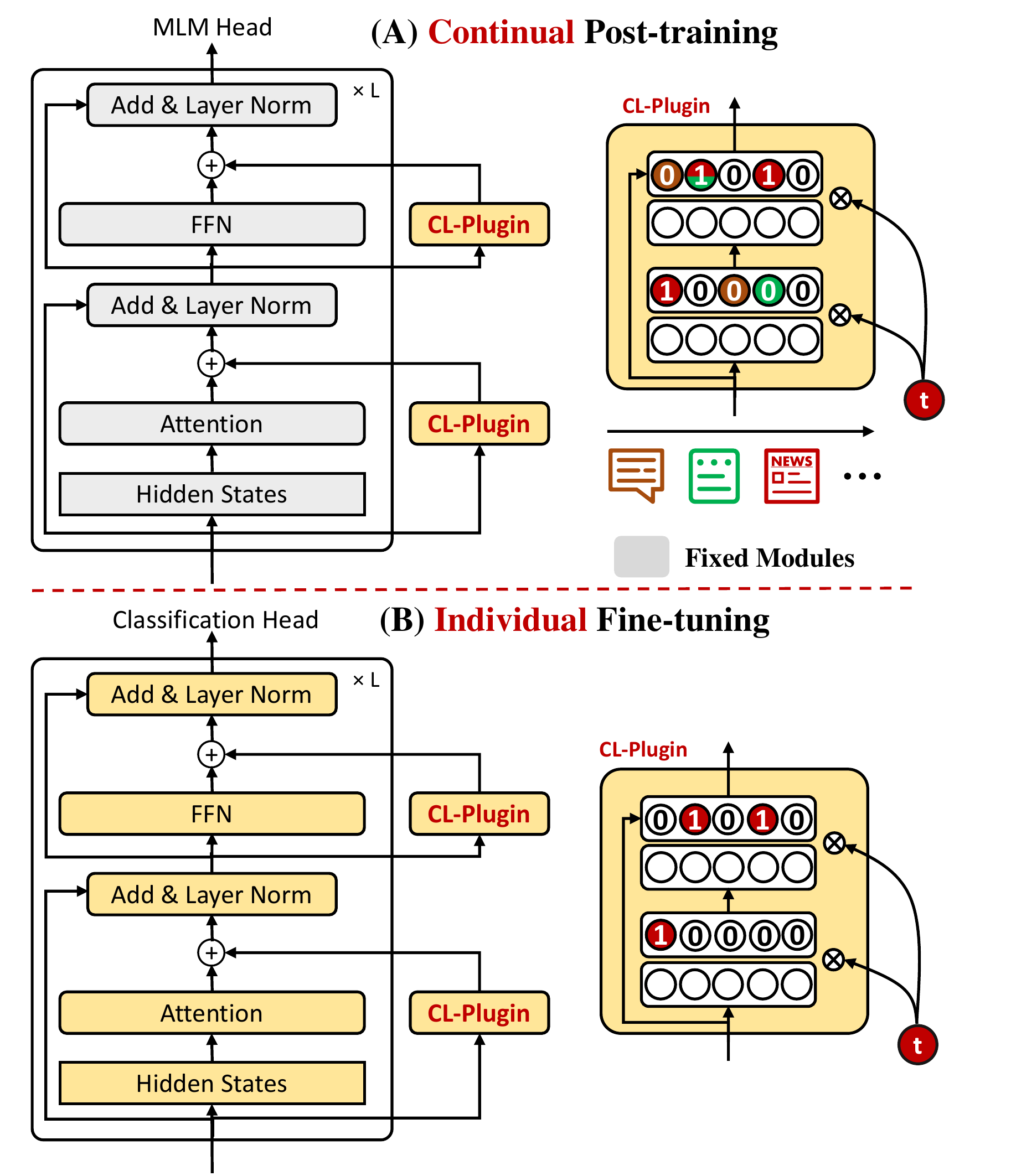}
\vspace{-6mm}
\caption{
%\zixuan{TODO: remove the download link. What are the grey and green color in Transformer means? (Originally they are to indicate trainability. NOw we have two things indicate trainability, which is confusing. Update the one in appedix as well}
Architecture of CPT, which has two CL-plugins inserted in the transformer layers of RoBERTa in a parallel manner (FFN: feed-forward network). \textbf{(A)} CPT for \textit{continual post-training}. It uses a masked language model (MLM) head for unsupervised post-training of the CL-plugins only. \textbf{(B)} CPT for \textit{individual fine-tuning}. CPT is evaluated by the corresponding individual end-task performance of all post-trained tasks. 
% The end-task user applies \bing{no need to say download. Remove the download link} CPT for his/her end-task and fine-tunes the whole model with the CL-plugins using a supervised classification head.} 
%  \bing{not described in text}
Each CL-plugin has numbers and colors indicating its masks and is illustrated in {Appendix} \ref{sec.illustration}.
}
\label{overview}
%\vspace{-4mm}
\end{figure}

CPT continually post-trains RoBERTa~\cite{DBLP:journals/corr/abs-1907-11692}. This is achieved by two \textit{continual learning plug-in} (called CL-plugin) modules inserted into each transformer layer of RoBERTa. CL-plugin is inspired by adapters in~\cite{Houlsby2019Parameter}. While adapters can isolate different tasks, one needs to allocate a new adapter for each task and no knowledge can be shared among different tasks' adapters. The CL-plugin, however, is a CL system that learns a sequence of tasks with adapters shared by all domains. 
% while an adapter adapts RoBERTa to one specific task (thus not a CL system). \zixuan{What is the advantage of CL-plugin, compared to Adapter-ONE, if there is no transfer.}
Figure~\ref{overview} gives the CPT architecture with two CL-plugins added to RoBERTa. 

{\color{black}\textbf{Sequential vs. Parallel CL-plugin.} Instead of following the original sequential adapter \cite{Houlsby2019Parameter}, CL-plugin adopts the parallel adapter idea in \cite{DBLP:journals/corr/abs-2110-04366}. The difference is that the former inserts an adapter \textit{after} the FFN/attention layer while the latter inserts it \textit{before} FFN/attention layer (see Fig. \ref{overview}). We choose the parallel version as it performs better (see Sec. \ref{sec:results}). 
}

In post-training, only the two CL-plugins are trained. The components of the original pre-trained RoBERTa are fixed. In end-task fine-tuning, all components are trainable. A CL-plugin is a two-layer fully connected network with a task mask mechanism. It takes two inputs: (1) hidden states $h^{(t)}$ from the feed-forward layer in a transformer layer and (2) task ID $t$ needed by task incremental learning (TIL). 
%The outputs are hidden states with features suitable for the $t$-th task for classification. 
Inside a CL-plugin, task masks (TMs), which indicate task- specific neurons, are used to {deal with CF}. Since TMs is differentiable, the whole CPT can be trained end-to-end. % We detail each module below. 

\subsection{Task Masks (TMs)} 
\label{sec:task_mask}
% \zixuan{if we go Machine Learning track, we may need some math}
%\zixuan{I put transfer here as well}

% \zixuan{I add "butterfly effect", not sure if it is okay}

% We now discuss how to preserve task-specific knowledge of previous tasks to prevent CF. % and leave more neurons for knowledge sharing.
%Besides knowledge sharing, we also need to capture the task-specific knowledge. 
%In learning the task specific knowledge, we want to 
%We want to ensure the task specific knowledge of old tasks that cannot be shared with the new task is not changed by the new task; otherwise sever catastrophic forgetting (CF) occurs.
 % to restrict the modification of neurons and keep such masks consistent throughout the entire CL process. 
% We use task masks (CL-plugin module in Figure~\ref{overview}). Specifically, in learning a new task, we first use masks learned for each task in the past to block off the neurons that are important for each old task. In the process of learning the new task, the masks are also learned for the task to be used later. 

In each layer of a CL-plugin, task masks are used to protect those neurons that are important for previous tasks to overcome CF. The masks basically forbid gradient updates to those neurons during backpropagation in learning a new task. Note that a task is also a domain in our case.

Learning a new task/domain consists of two main steps: (1) apply the mask in each layer for each old task to block off the gradient flow to protect the model for the old task, and (2) learn domain $t$ and its masks for future use. We present (2) first.  
 
%\hu{to encourage knowledge sharing}.
%\hu{check if we can claim this or not} 
% \zixuan{We need to decide how detail we want, Or we just say: "Here we borrow the hard attention idea in \cite{Serra2018overcoming} and leverage the task ID embedding to train the task mask. Further details can be found in Appendix." as in CPT} 

%\zixuan{I changed bellow for as a simplified version}
% \zixuan{I commented out all the illustration, and put them in the caption of Fig. 1}
% finds those neurons are also useful. % and supplies those neurons with shareable knowledge from KSM. 
% This allows for \textit{backward transfer}.

% \iffalse

% \textbf{Task Masks.}~Given the transfer capsule $v_j^{(t)}$, TSM maps them into input $k_l^{(t)}$ via a fully-connected network, where $l$ is the $l$-th layer in TSM. 
%During the mapping, the task-specific module aims to learn features 
%neurons that are used by the previous tasks (\bing{?? what tasks?}) and blocking the gradient flow through the neurons (i.e., setting their gradients to 0). 

%\noindent \textbf{From Task ID Embedding to Task Mask.}
%\textbf{From Task ID to Task Specific Module. } 
% \hu{I moved from a previous paragraph.}

\textbf{Learning Task Masks for Overcoming CF.} In learning each task $t$, a mask (a ``soft'' binary mask) ${m}^{(t)}_l$ is trained for the task at each layer $l$ in CL-plugin, indicating the neurons that are important for the task. We borrow the hard attention idea in \cite{Serra2018overcoming} and leverage the task ID embedding to train the mask. For a task ID $t$, its embedding $e^{(t)}_l$ consists of differentiable deterministic parameters that can be learned together with other parts of the network. 
%The subscript $l$ indicates the layer number. 
%A separate task ID embedding 
% It is trained for each layer in CL-plugin.
To generate the task mask $\text{m}^{(t)}_l$ from $e^{(t)}_l$, \textit{Sigmoid} is used as a pseudo-gate (mask) function. % \zixuan{I removed the s. since it has been changed to $\tau$}
% and a positive %\hu{the formula seems can take negative too} 
% scaling hyper-parameter $s$ is applied to help training. 
$m^{(t)}_l$ is computed with
%\vspace{-1.5mm}
\begin{equation}
% \vspace{-1.5mm}
\label{eq:mask}
m^{(t)}_l = \sigma(e^{(t)}_l/\tau),
\end{equation}
where $\tau$ is a temperature variable, linearly annealed from 1 to $\tau_{\min}$ (a small positive value). 

In the forward pass, given the output of each layer $l$, $k_l^{(t)}$, we element-wise multiply mask $m^{(t)}_l$,
\begin{equation}
\label{eq:forward}
o_l^{(t)} = k_l^{(t)} \otimes m^{(t)}_l.
\end{equation}
{The masked output $o_l^{(t)}$ of the last layer in CL-plugin is} fed to the next layer of the RoBERTa with a skip-connection. After learning task $t$, the final $m^{(t)}_l$ is saved and added to the set $\{m^{(t)}_l\}$.

\textbf{Applying Task Masks.}
 %$i_{\text{prev}}$. 
Before learning a new task $t$, we first accumulate and set the masks $m^{(i_{\text{prev}})}_l$ on the neurons in each layer $l$ for all old tasks $i_{\text{prev}}$
so that in backpropagation, the gradient $g^{(t)}_l$ for task $t$ will not flow to these neurons. 
Since $m^{(i_{\text{prev}})}_l$ is {pseudo} binary, we use 
max-pooling to achieve the accumulation and condition the gradient:  
%\vspace{-1.5mm}
% \begin{equation}
% \label{eq:accumulate}
% m^{(t_{\text{a}})}_l = \.
% \end{equation}
% \hu{it's better to replace tm to other variable: m?}
% \begin{equation}
% m^{(\leq i_{dis})}_l = max(m^{(i_{dis})}_l, m^{(\leq (r_{dis}^{(i-1)})}_l)
% \end{equation}
% The term $m^{(t_{\text{a}})}_l$ is applied to the gradient:
% \vspace{-1.5mm}
\begin{equation}
\label{eq:gradient}
g^{'(t)}_l = g^{(t)}_l \otimes (1-(\text{MaxPool}(\{m^{(i_{\text{prev}})}_l\}))).
\end{equation}
Those gradients corresponding to the 1 entries in $\text{MaxPool}(\{m^{(i_{\text{prev}})}_l\})$ are set to 0 (to block off gradient flow) while the others remain unchanged. 
In this way, neurons in old tasks are protected. Note that we expand (copy) the vector $m_l^{(t_{\text{a}})}$ to match the dimensions of $g_l^{(t)}$.

\subsection{Catastrophic Butterfly Effect in Fine-tuning}
\label{sec:butterfly}
% \zixuan{I noticed the content about pusedo-mask in fine-tuning is removed. I feel only make the mask real binary in gradient mannipulation during DA-training is not enough. During fine-tunning, we also need to "select" neurons and we need real binary mask to achieve perfect selection.} \bing{please add} \zixuan{changed}

To perform an end-task in a post-trained domain, we fine-tune the mask-protected model of the domain, which is indicated by the task/domain id. The fine-tuning uses the corresponding domain neurons for the specific end-task {by setting $\tau=\tau_{\text{min}}$ and condition the output via Eq. \ref{eq:forward}}. With the masks, there should be no forgetting for continual post-training and the end-task fine-tuning performance should be similar to post-train each domain separately. % regardless {of} how many domains have been learned during continual post-training. 
However, we found that this is not the case.\footnote{For example, fine-tuning an end restaurant sentiment classification task achieves macro-F1 (MF1) of 0.64 right after post-training the restaurant domain but its fine-tuning MF1 drops to 0.44 after post-training three more domains.} Our investigation found that the problem is due to the \textit{pseudo-gate} function in Eq.~\ref{eq:mask}. No matter how small $\tau$ is, Eq. \ref{eq:mask} can only gives us a mask almost 0 (or 1). This causes the following: (1) During post-training, 
% the gradient after Eq.~\ref{eq:mask} is not exactly 0 but a very small number.
{the gradients for used neurons in Eq.~\ref{eq:gradient} are not exactly 0 but a very small number.} (2) During fine-tuning, we cannot make use of the corresponding neurons for the specific end-task by simply setting $\tau=\tau_{\text{min}}$. % Even if we use the corresponding neurons for a specific end task via Eq. \ref{eq:forward} by setting $\tau=\tau_{\text{min}}$, the neurons for other tasks will still have a very small activation instead of exactly 0. 
The small change in the neurons for old domains during post-training caused by (1) is neglect-able in conventional CL because in conventional CL we evaluate the model using test sets and no weights update involved. However, in CPT, the end-task needs to fine-tune the continually post-trained LM model (p-LM), which involves weight updating. A small change to the p-LM during continual post-training can result in a different initialization for the end-task fine-tuning and give totally different fine-tuning results. We call this \textit{butterfly effect} inspired by the term indicating a small state change in  nature (e.g., the flap of a butterfly’s wings in Brazil) can result in large differences in a later state (e.g., a tornado in Texas). % we call our observation \textit{butterfly effect in fine-tuning}, where a small change in the previous post-trained task neurons can result in a large differences after fine-tuning. The butterfly effect in fine-tuning tells us we need to have a strict protection method during post-training and requires us to have a \textit{hard binary} mask to make sure the forgetting will not happen.

We propose a simple method to solve it, i.e., adding a threshold $\theta$ to the $m^{(t)}_l$ to make it a \textit{hard binary mask},
\begin{equation}
\label{eq:threshold}
m^{(t)}_l =\left\{
\begin{aligned}
1 & , & m^{(t)}_l > \theta, \\
0 & , & m^{(t)}_l < \theta.
\end{aligned}
\right.
\end{equation}
%\zixuan{I changed Eq. 4. if m > 0.5, it should be 1}
We then apply it to Eq. \ref{eq:gradient} in gradient manipulation and Eq. \ref{eq:forward} in end-task fine-tuning. $\theta$ can be easily set (we use 0.5) since Eq.~\ref{eq:mask} already gives a pseudo-binary mask. Note that this has almost no effect on post-training % in Eq.~\ref{eq:mask} and Eq.~\ref{eq:forward}. 
as it is used to block the backward pass gradient flow during post-training and select the corresponding neurons during fine-tuning. 

% \zixuan{Implementation is moved to appendix}
% \subsection{Implementation Detail}
% \textbf{Architecture.} A masked language model head is applied to for post-training and a fully connected layer with softmax output is used as the classification heads in the last layer of the RoBERTa for fine-tuning, together with the categorical cross-entropy loss.

% \textbf{Hyperparameters.}
% Unless otherwise stated, the same hyper-parameters are used in all experiments.  We use $0.0025$ for $\tau_{\min}$ in Eq.~\ref{eq:mask} and $\theta$ is set to 0.5 in Eq.~\ref{eq:threshold}. We use 2000 dimensions as the final and the hidden layers of the CL-plugin. The task ID embeddings have 2000 dimensions.  The training of RoBERTa, Adapter-RoBERTa and CPT follows that of \cite{DBLP:conf/naacl/XuLSY19}. We adopt $\text{RoBERTa}_{\textbf{BASE}}$ (uncased). The maximum input length is set to 128 which is long enough for all datasets. We use Adam optimizer and set the learning rate to 1e-4 for post-training and 1e-5 for fine-tuning. The batch size is set to 128 for post-training and 32 for fine-tuning. We train CPT on each task/domain for 13K steps for post-trainng, which is approximately a single pass on each domain dataset. We train on fine-tuning datasets for 30 epochs and take the results for the last epochs. 

\begin{table*}[]
\centering
\resizebox{0.8\textwidth}{!}{
\begin{tabular}{cc|cccccccccccc}
\specialrule{.2em}{.1em}{.1em}
\multirow{2}{*}{Category} & Domain & \multicolumn{2}{c}{Restaurant} & \multicolumn{2}{c}{AI} & \multicolumn{2}{c}{ACL} & \multicolumn{2}{c}{AGNews} & \multicolumn{2}{c}{Average} & \multicolumn{2}{c}{Forget R.} \\
 & Model & MF1 & Acc & MF1 & Acc & MF1 & Acc & MF1 & Acc & MF1 & Acc & MF1 & Acc \\
\specialrule{.1em}{.05em}{.05em}
\multirow{4}{*}{Non-CL} & RoBERTa & 50.61 & 74.77 & 27.88 & 28.44 & 32.19 & 34.59 & 64.19 & 65.95 & 43.72 & 50.94 & \multicolumn{2}{c}{---} \\
& Adapter & 45.40 & 67.28 & 23.69 & 24.56 & 24.99 & 27.55 & \textbf{64.53} & \textbf{66.50} & 39.65 & 46.48 & \multicolumn{2}{c}{---} \\
 & RoBERTa-ONE & 53.63 & \textbf{76.73} & 29.86 & 30.11 & 33.05 & 35.72 & 62.57 & 65.13 & 44.78 &51.92 & \multicolumn{2}{c}{---} \\
 & Adapter-ONE & 52.19 & 74.20 & 30.80 & 31.59 &36.59 & 36.99 & 61.66 & 63.94 & 45.31 & 51.68 & \multicolumn{2}{c}{---} \\
 & Prompt-ONE & 28.93 & 59.79 & 21.06 & 22.10 & 28.02 & 29.22 & 60.70 & 62.58 & 34.68 & 43.42 &
 \multicolumn{2}{c}{---} \\
 & DEMIX & 53.14 & 75.28 & 27.68 & 27.29 & 37.63 & 38.57 & 63.18 & 65.13 & 45.41 & 51.57 &
 \multicolumn{2}{c}{---} \\
\hline
\multirow{8}{*}{CL} & RoBERTa-NCL & 42.59 & 67.56 & \textbf{31.57} & \textbf{31.62} & 33.07 & 34.54 & 60.18 & 63.50 & 41.85 & 49.30 & 3.27 & 2.82 \\
 & Adapter-NCL & 47.42 & 70.23 & 29.56 & 29.90 & 35.92 & 37.58 & 61.73 & 64.45 & 43.65 & 50.54 & 2.21 & 1.69 \\
 & HAT & 50.45 & 71.78 & 28.33 & 29.41 & 34.93 & 37.15 & 62.97 & 65.05 & 44.17 & 50.85 & 2.43 & 2.04 \\
%& CLASSIC & 50.39 & 72.61 & 29.28 & 29.59 & 35.16 & 35.10 & 64.03 & 66.63 & 44.71 & 50.98 & 1.85 & 1.93 \\
& BCL & 51.70 & 74.34 & 29.66 & 30.96 & 32.85 & 34.82 & 63.60 & 65.47 & 44.45 & 51.40 & 1.47 & 0.82 \\
 & KD & 39.75 & 67.11 & 29.63 & 29.33 & \textbf{38.30} & \textbf{42.09} & 62.85 & 65.39 & 42.63 & 50.98 & 4.92 & 3.07 \\
 & EWC & 48.32 & 71.59 & 30.96 & 31.01 & 35.96 & 38.05 & 62.29 & 64.95 & 44.38 & 51.40 & 1.40 & 0.80 \\
& DER++ & 48.09 & 71.79 & 30.71& 30.54 &34.25 & 35.77 & 64.24 & 66.11 & 44.32 & 51.05 & 1.79 & 1.62 \\
 & CPT & \textbf{53.90} & 75.13 & 30.42 & 30.89 & 37.56 & 38.53 & 63.77 & 65.79 & \textbf{46.41} & \textbf{52.59} & \textbf{0.00} & \textbf{0.00} \\
\specialrule{.1em}{.05em}{.05em}
\end{tabular}}
\caption{End-task macro-F1 (MF1), accuracy and forgetting rate results 
% (Sec.~\ref{sec:results}) 
for all domains \textit{after continual post-training of all domains}. The results are averages of 5 random seeds (the domain training order is as they appear in the first row). Due to space limits, the results for \textit{different domain orders} and the \textit{standard deviations} are reported in Appendix~\ref{ap:order} and Appendix~\ref{ap:std}, respectively). Non-CL baselines has no forgetting. 
% \zixuan{Do we want to add TAPT results--- we are not better than MultiTask (MLM+CE) setting. Camera domain  checked}
} 
\label{tab:overall_results}
% \vspace{-4mm}
\end{table*}

\section{Experiments}
\label{Sectionexperiments}
% We evaluate CPT using \textbf{4} domain adaptive datasets and \textbf{4} text classification end task. 
% We follow the continual learning (CL) evaluation method given in~\citep{DBLP:journals/corr/abs-1909-08383}.
% \hu{not sure if ICML reviewers already know these well (seems totally not a problem for naacl reviewers in sentiment).}
The proposed paradigm uses a different evaluation from that of conventional continual learning (CL). After unsupervised continual post-training of an LM (RoBERTa in our case) with a sequence of domains, the resulting p-LM is used to fine-tune an end {\color{black}few-shot classification task} from any post-trained domain. There is no CL during end-task fine-tuning. Each fine-tuning task is done separately. % can neither affect each other nor the post-training model. %\hu{do we need to breakdown clearly how to evaluate back/forward transfer and forgetting here?} \hu{is this sentence obvious (in CL)?}
% In training each task, we use its validation set to decide when to stop training. 

\subsection{Datasets and Baselines}
\label{sec.data-baselines}
% \zixuan{Introduce our own datasets, restaurant	- laptop - 	agnews - 	yahoo - 	dbpedia -	mnli}
\textbf{Datasets:} We use 4 \textbf{unlabeled domain datasets}: \textit{Yelp Restaurant} \cite{DBLP:conf/naacl/XuLSY19}, \textit{AI Papers} \cite{DBLP:conf/acl/LoWNKW20}, \textit{ACL Papers} \cite{DBLP:conf/acl/LoWNKW20} and \textit{AGNews} \cite{DBLP:conf/nips/ZhangZL15} and their 4 corresponding \textbf{end-task classification datasets}.\footnote{These are popularly used in related works. Details of the datasets are given in {Appendix} \ref{sec:data_stat}.
We conduct experiments using \textit{few-shot} learning end-tasks. Following~\cite{gu2021ppt}, we use 32 training samples for \textit{Restaurant} and \textit{AGNews}, 48 training samples for \textit{ACL} and 56 training samples for \textit{AI} due to different numbers of classes in each dataset.}
% and ensure the number of labels is balanced for end-tasks with less than 5 labels. For end-tasks with more than 5 labels, we randomly select 8 training samples for each label. }

% \zixuan{Datasets section moved to Appendix} \bing{we still need to list the datasets here and only leave the detail to appendix}

% The first two applications (and datasets) are used to show the knowledge transferability of CPT because their tasks are similar and have shared knowledge. Catastrophic forgetting (CF) is not a major concern for them. The third application (and dataset) is mainly used to test CPT's ability to overcome CF as its tasks are very different and have little shared knowledge to transfer.  

\vspace{+1.5mm}
\noindent
\textbf{Baselines.} Since no existing method can perform our task, we use 6 \textit{non-CL} and 7 \textit{adapted CL} methods as our baselines. The non-CL baselines include \textbf{(1) RoBERTa} and \textbf{(2) Adapter} where we directly fine-tune the pre-trained model or adpater (without any post-training); \textbf{(3) RoBERTa-ONE}, \textbf{(4) Adapter-ONE} and \textbf{(5) Prompt-ONE}, where we build a model for each task using a separate network. It has no knowledge transfer or CF. \textbf{(6) DEMIX} \cite{gururangan2021demix} trains a separate adapter for each task and initializes the adapter from its most similar previous task adapter. The 7 adapted CL baselines include \textbf{(7) RoBERTa-NCL} and \textbf{(8) Adapter-NCL}, where we post-train the domains one by one with no mechanism to deal with CF/transfer. Other are state-of-the-art CL baselines and we adapt them for continual post-training.\footnote{Readers can refer to Appendix \ref{ap:baselines} for the detailed of each of these baselines.}

{\color{black}\subsection{Implementation Details} %They are given in Appendix~\ref{sec:imp_detail}. 
\textbf{Architecture.} We adopt $\text{RoBERTa}_{\textbf{BASE}}$ as our backbone LM. A masked language model head is applied for post-training. The fine-tuning follows the standard practice \cite{DBLP:conf/naacl/DevlinCLT19}, where we pass the final layer \texttt{</s>} token representation to a task-specific feed-forward layer for prediction. The feed-forward layer with softmax output is used as the classification heads, together with the categorical cross-entropy loss. Note that for the aspect sentiment classification task (see Table~\ref{tab:dataset}), we adopt the ASC formulation in \cite{DBLP:conf/naacl/XuLSY19}, where the aspect (e.g., ``\textit{sound}'') and review sentence (e.g., ``\textit{The sound is great}'') are concatenated via \texttt{</s>}. 

\textbf{Hyperparameters.}
Unless otherwise stated, the same hyper-parameters are used in all experiments.  We use $0.0025$ for $\tau_{\min}$ in Eq.~\ref{eq:mask} and $\theta$ is set to 0.5 in Eq.~\ref{eq:threshold} in the main paper. 
% Following~\cite{jang2021continual}, we set ffn\_size\footnote{dimensions for the hidden layer of CL-plugin after the FFN sub-layer.} larger than attn\_size\footnote{dimensions for the hidden layer of CL-plugin after the multi-head attention.}. 
As shown in Figure \ref{overview}, there are two CL-plugins for each Transformer layer (one at the bottom in parallel with attention and one at the top in parallel with FFN). We search the CL-plugin size within \{128, 256, 512, 768, 1024\} and adopt 512 for the bottom one and 768 for the top one based on validation experiments. The task id embeddings have the same size as the hidden layer dimension of the CL-plugin.
The maximum input length is set to 164 which is long enough for all datasets. We use Adam optimizer and set the learning rate to 1e-4 for post-training and 5e-5 for fine-tuning. The batch size is set to 48 for post-training and 20 for fine-tuning. Since each of our domain-specific dataset has a different size, we train CPT on each task/domain for 1 epoch for post-training, which is approximately 13K steps, following \cite{DBLP:conf/acl/GururanganMSLBD20,DBLP:conf/naacl/XuLSY19}. We train on end-task fine-tuning datasets for 20 epochs and take the results for the last epoch, assuming no validation sets. We empirically found 20 epochs can give us a relatively stable results.} 
\subsection{Evaluation Results and Analysis}
% \zixuan{need update}
\label{sec:results}
% \zixuan{we may not have these random sequence}
% Since the order of the tasks in a sequence may impact the final results, we randomly sampled 5 task sequences and averaged their results.
We report the average results of the 4 different fine-tuning tasks (or datasets) in accuracy and Macro-F1 after post-training on all unlabeled domain datasets in
Table~\ref{tab:overall_results}. The forgetting rate (forget R.) \cite{DBLP:conf/cvpr/LiuSLSS20} is also reported. The higher the forgetting rate is, the more forgetting it has. Negative rates indicate positive knowledge transfer.\footnote{Forgetting rate is {\color{black}computed as as follows~\cite{DBLP:conf/cvpr/LiuSLSS20},} $\frac{1}{t-1}\sum_{i=1}^{t-1}A_{i,i} - A_{t,i}$, where $A_{i,i}$ is the end-task performance right after its domain $i$ is post-trained, and $A_{t,i}$ is the performance of the end-task of domain $i$ after post-training the last domain. We average over all end-tasks except the last one as the last domain has no forgetting.}
% \textbf{Overall Performance.}
% \ls{I feel confused in the baseline name. the baseline composed with pretrainmodule+pretraintype+finetunemodule. what does Comb refer? Please aware that RoBERTa+Comb+plugin performance is not bad. CPT is not significant better than RoBERTa+Comb+plugin.}\zixuan{changed the table}
% which shows that 
% The detailed results for each dataset are given in \textit{Appendix}

\textbf{Superiority of CPT.} Clearly, CPT outperforms all baselines and achieves no forgetting. More specifically, CPT markedly outperforms the two baselines without post-training (RoBERTa and Adapter), indicating CPT can learn new domains well. These two baselines are also  significantly worse than other baselines, indicating that fine-tuning the pre-trained RoBERTa alone is weak. 
% performs poorly in domains {\color{red}?? outside its respective training domain}, as mentioned in the Introduction. 
Comparing with CL baselines, \textbf{CPT achieves \textit{no forgetting}} (we can see the forgetting rate is 0), indicating the high effectiveness of the proposed approach. We also note that CPT is even slightly better than those ONE baselines, indicating \textbf{\textit{some positive knowledge transfer in CPT}}. 
%  \bing{not defined}
 
\subsection{Ablations} 
In Table \ref{tab:ablation_results}, we give the ablation results. We are interested in the following:

% \noindent
\textbf{(1) Catastrophic butterfly effect} (CBE). The third row with ``w/o butterfly'' shows results without the hard binary mask mechanism in Eq.~\ref{eq:threshold}. Clearly, the results are worse and the model suffers from forgetting. This indicates CBE and our approach is effective. 

\textbf{(2) Different Architecture.} CPT is based on CL-plugin, which is inspired by adapters. Another popular way to use adapters is to make it sequential \cite{Houlsby2019Parameter}. Sequential adapter (first row) is clearly poorer than our current parallel one. This conforms to the observation in \cite{DBLP:journals/corr/abs-2110-04366}.

\textbf{(3) Different Orders.} Table \ref{tab:overall_results} only reports the results of one fixed domain order (\texttt{Restaurant}$\to$\texttt{AI}$\to\texttt{ACL}\to$\texttt{AGNews}). We are interested in how the order impacts CPT results. We give the detailed results for all the other baselines and detailed domain orders in Appendix~\ref{ap:order}. We can see the results of CPT does not change much and it still outperforms other baselines. This indicates the CPT's robustness to domain orders in  post-training.

\begin{table}[]
\centering
\resizebox{0.7\columnwidth}{!}{
\begin{tabular}{c||ccc}
\specialrule{.2em}{.1em}{.1em}
\multirow{2}{*}{Model} & \multicolumn{2}{c}{Final Performance}  \\
 & MF1 & Acc \\
\specialrule{.1em}{.05em}{.05em}
CPT (Sequential Adapter)  &  43.00  & 50.25  \\
\hline
CPT (w/o butterfly)  &  44.17  & 50.85   \\
CPT (w/o masking)  &  43.65  & 50.54   \\
% \hline
% CPT (sequence 1)  &  46.49  & 52.47   \\
% CPT (seqeence 2)  &  45.71  & 51.71   \\
% CPT (seqeence 3)  &  46.15  & 51.93   \\
% CPT (seqeence 4)  &  45.89  & 51.86   \\
\hline
CPT  & 46.41 &  52.58 \\
\specialrule{.1em}{.05em}{.05em}
%\vspace{-4mm}
\end{tabular}
}
% \vspace{-2mm}
\caption{Ablation experiment results. }
\label{tab:ablation_results}
\vspace{-6mm}
\end{table}

%\subsection{Error Analysis}
% \hu{how some patterns of sequence of tasks, break down your 5 random and see why some are worse because of some ordering? visualization of masks with domain labels, and capsule hidden space? } \zixuan{Do we need it? }

\section{Conclusion}
{\color{black}This paper proposed to continually post-train an LM with a sequence of domains using their unlabeled domain corpora. An effective method (CPT) is also proposed. An end-task from any post-trained domain can fine-tune the resulting LM. Experimental results demonstrate the effectiveness of CPT. % performs significantly better in few-shot learning than without continual domain-adaptive pre-training.  % studied task continual learning (Task-CL) using the pre-trained model RoBERTa to achieve both CF presentation and knowledge transfer.
} 
% It proposed a novel technique called CPT to leverage the pre-trained RoBERTa for CL. The key component of CPT is the CL-plugin inserted in RoBERTa. A CL-plugin is a capsule network with a new transfer routing mechanism to encourage knowledge transfer among tasks and also to isolate task-specific knowledge to avoid forgetting. Experimental results using three NLP applications showed that CPT markedly improves the performance of both the new task and the old tasks via knowledge transfer and is also effective at overcoming catastrophic forgetting.
% {\color{black}One limitation of our work is the efficiency due to the use of capsules. Capsules try to represent a group of neurons in a vector reflecting properties of an entity. In NLP, an entity is a sentence/document which contains many tokens (e.g., 128) and features (e.g. 768 in $\text{RoBERTa}_{\text{BASE}}$). Grouping them makes the capsule very large (e.g., 128 $\times$ 768), which slows down training.}

\section{Limitations}
{\color{black}We list two limitations of CPT. First, CPT adds CL-plugins for continual post-training with no change to the underlying LM in training. Although a CL-plugin is small compared to an LM, it is still interesting and may be more effective to explore the idea of updating the LM directly without any additional modules. Second, domain ids are needed in both training and testing for CPT. In some applications, it may be hard to provide a domain id for each fine-tuning end-task. We will explore these in our future work as specializing and/or incrementally improving an LM is an important problem.} % relies on a \textit{subset} of domain data. If the subset is not representative enough, DGA may not be able to find the correct importance. \bing{why cannot we use the full domain data?}\zixuan{because full domain data is huge. One typiucally don't want to post-trainin the data twice}

\section*{Acknowledgments}
{\color{black}The work of Zixuan Ke and Bing Liu was supported in part by two National Science Foundation (NSF) grants (IIS-1910424 and IIS-1838770).} 

\bibliography{anthology,custom}
\bibliographystyle{acl_natbib}

\appendix

\newpage

% \vspace{-2mm}
\section{Related Work}
\label{Sectionrelated.work}
% \zixuan{Section 2 are copied from ECML paper}
% As we are not aware of any reported work that uses RoBERTa fine-tuning and capsules in CL, here we discuss the related work in CL in general with regard to overcoming CF and knowledge transfer. % and then some specific CL approaches for solving the problems that use in our experiments. \hu{$<-$remove?}
% \ls{I think DEMIX is a baseline. At leaset, we may need to discuss it. In DeMIX, it claimed continual domain LM and non-forgetting.}

% \ls{do we have concept-level difference to them? detailed techniques and evaluation are too fine-grain. reviewer may request exp on demix for fine-tuning.}
% \zixuan{need mentioned the difference with few-shot learning, we  may need to deemphasize the knwoeldge transfer}

Our work is related to \textit{continual learning}, \textit{post-training} and \textit{few-shot learning}.

\textbf{Continual learning (CL).} In general, overcoming CF is a major goal in CL~\cite{chen2018lifelong}. % There are several methods to achieve this. 
(1) \textit{Regularization methods}~\cite{Kirkpatrick2017overcoming,Seff2017continual} add a regularization to ensure minimal changes to weights for previous tasks.
(2) \textit{Replay methods} retain~\cite{Rebuffi2017,Lopez2017gradient,wang2020efficient,guo2022online} or generate some data of old tasks~\cite{Shin2017continual,He2018overcoming} and use them in learning a new task.
(3) \textit{Parameter isolation methods} \cite{Serra2018overcoming,fernando2017pathnet} allocate parameters for different tasks and mask them out in learning a new task. {\color{black}Our CPT is based on (3) and uses masks in continual post-training.} Recently, CL has drawn attention in NLP. It has been used for slot filling~\cite{shen-etal-2019-progressive}, language learning~\cite{li2019compositional}, sentence embedding~\cite{liu2019continual}, translation~\cite{khayrallah2018regularized}, cross-lingual modeling~\cite{liu2020exploring}, question answering~\cite{greco2019psycholinguistics} and text classification~\cite{DBLP:journals/corr/abs-2112-02706,ke2021Classic,sun2020lamol,huang2021continual,chuang2020lifelong,mehta2021empirical,madotto2020continual}. However, none of them tries to improve an LM.

\textbf{Post-training} is an effective approach to mitigate the discrepancies between pre-trained domains and the target domain. Researchers have applied post-training to many domains, e.g., reviews \cite{DBLP:conf/naacl/XuLSY19,sun2019fine}, news and academic papers \cite{DBLP:conf/acl/GururanganMSLBD20}, and shown improved end-task results. However, none of them consider the continual learning paradigm. 

\textbf{Few-shot learning (FL)} aims to learn tasks with a few labeled examples. The main issue of FL is over-fitting, due to the scarcity of
labeled training data. Existing methods to overcome over-fitting fall in three main families: (i) model-based methods try to reduce the hypothesis space of the few-shot task \cite{DBLP:conf/nips/TriantafillouZU17,DBLP:conf/coling/HuLT0S18}, (ii) data-based methods try to augment additional data to the few-shot set \cite{DBLP:conf/nips/BenaimW18,DBLP:conf/aaai/GaoHX0LLS20}, and (iii) algorithm-based solutions try to improve strategies for searching for the best hypothesis. Recently, a new
paradigm using prompts achieves promising results for few-shot language learning as shown
in GPT-3 \cite{brown2020language}, PET \cite{DBLP:conf/eacl/SchickS21} and LM-BFF \cite{DBLP:conf/acl/GaoFC20}. {\color{black}However, none of them does few-shot fine-tuning in continual post-training. 

\textbf{Continual few-shot learning.} Several researchers have studied this problem recently \cite{DBLP:journals/corr/abs-2004-11967,DBLP:journals/corr/abs-2110-07298,DBLP:conf/emnlp/JinLR021,DBLP:conf/naacl/XiaYFY21,DBLP:conf/acl/0001LX22}. It continually learns a sequence of few-shot tasks. However, this is very different from our continual post-training because our continual learning happens in the post-training stage instead of the end-task fine-tuning stage. We only evaluate the proposed CPT system after continual post-training by conducting few-shot learning tasks individually by fine-tuning the post-trained language model (p-LM) in each of the post-trained domains. No continual learning is involved in few-shot learning.}
% \bing{any few-shot continual learning methods?}

\begin{figure*}[h]
\centering
\includegraphics[width=\textwidth]{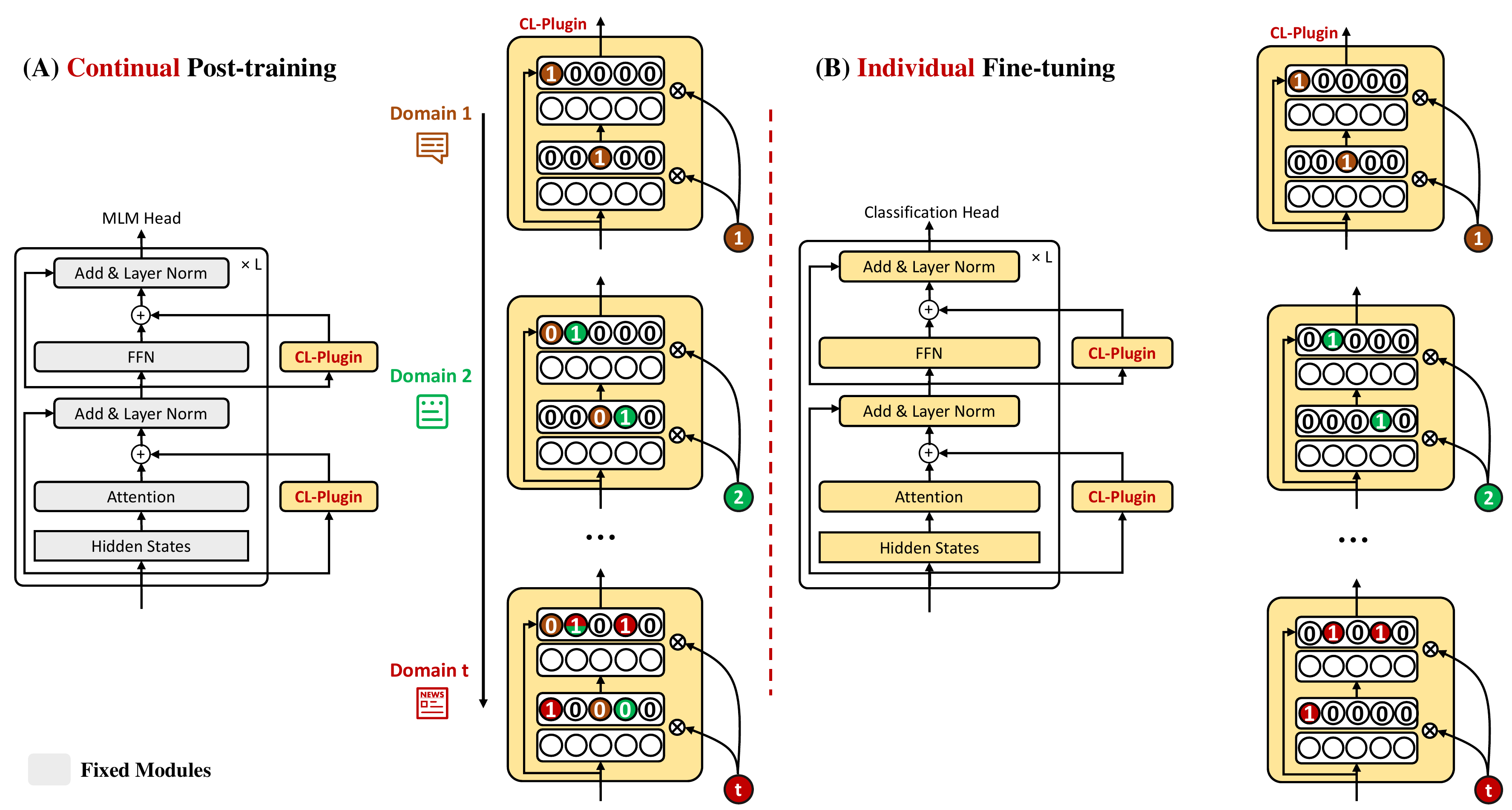}
\caption{
% \hu{you can make rays out of the lower CL-plugin, too.} 
% \hu{heard from Prof and Zixuan's suggestion: what is a good name for the whole pic? Adaformer, CLformer, Transformer-CL? any other ideas?} 
% Adapter-RoBERTa in \cite{DBLP:conf/icml/HoulsbyGJMLGAG19} and our RoBERTa-based continual learning (BACK). 
% \zixuan{need to change figure} 
Architecture of CPT, which has two CL-plugins inserted in the transformer layers of RoBERTa in a parallel manner. \textbf{(A)} CPT for \textit{continual post-training}. It uses a masked language model (MLM) head for unsupervised post-training of the plugins only. \textbf{(B)} CPT for \textit{individual fine-tuning}. The performance of CPT is evaluated by the corresponding individual end-task performance of all post-trained tasks using the \textit{final} post-trained model (with different mask). Each CL-plugin module (\textbf{to the right of the transformer}) has two fully connected layers and a skip connection. On top of each fully connected layer, there is a mask computed from task ID $t$ with the same size as the fully connected layer.}
\label{plugin}
% \vspace{-4mm}
\end{figure*}

\begin{table*}[]
\centering
\resizebox{\textwidth}{!}{
\begin{tabular}{ccc|ccccc}
\specialrule{.2em}{.1em}{.1em}
\multicolumn{3}{c|}{Unlabeled Domain Datasets} & \multicolumn{4}{c}{End-Task Classification Datasets} \\
Dataset & Source & \#training & Dataset & Task & \#training & \#testing & \#classes \\
\specialrule{.1em}{.05em}{.05em}
Yelp Restaurant & Yelp Review & 1,132,359 & SemEval-res & Aspect Sentiment Classification & 32 & 1,120 & 3 \\
AI & AI Papers & 707,368 & SCIERC  & Relation   Classification & 56 & 2,388 & 7 \\
ACL & ACL Papers & 1,208,449 & ACL-ARC & Citation Intent Classification & 48 & 421 & 6 \\
AGNews & News Article & 73,750 & AGNews-FT & News  Classification & 32 & 7,568 & 4 \\
% DBPedia & Wikipedia & 560,000 & DBPedia & Ontology   Classification & 70,000 & 14 \\
% \multirow{2}{*}{MNLI} & \multirow{2}{*}{\begin{tabular}[c]{@{}c@{}}government, telephone, \\ fiction, travel, slate\end{tabular}} & \multirow{2}{*}{433,000} & BANKING77 & \multirow{2}{*}{Intent   Classification} & 15000 & 150 \\
%  &  &  & CLINIC150 &  & 13083 & 77 \\
\specialrule{.1em}{.05em}{.05em}
\end{tabular}
}
\caption{
Statistics for unlabeled domain datasets and end task supervised classification datasets. % 
} %\ls{DSC waste too much space}}
\label{tab:dataset}
\end{table*}

\begin{table*}
\centering
\resizebox{\textwidth}{!}{
\begin{tabular}{cccccccccccc} 
\specialrule{.2em}{.1em}{.1em}
\multirow{2}{*}{Category} & Domain      & \multicolumn{2}{c}{Restaurant} & \multicolumn{2}{c}{AI} & \multicolumn{2}{c}{ACL} & \multicolumn{2}{c}{AGNews} & \multicolumn{2}{c}{Average}  \\
                          & Model       & MF1    & Acc                   & MF1    & Acc           & MF1    & Acc            & MF1    & Acc               & MF1    & Acc                 \\ 
\specialrule{.1em}{.05em}{.05em}
\multirow{6}{*}{Non-CL}   & RoBERTa     & $\pm$0.0456 & $\pm$0.0274                & $\pm$0.0208 & $\pm$0.0233        & $\pm$0.0391 & $\pm$0.0338         & $\pm$0.0121 & $\pm$0.014             & $\pm$0.0066 & $\pm$0.0062              \\
                          & Adapter     & $\pm$0.0214 & $\pm$0.0223                & $\pm$0.0111 & $\pm$0.0102        & $\pm$0.0375 & $\pm$0.0386         & $\pm$0.0221 & $\pm$0.0224            & $\pm$0.0155 & $\pm$0.0142              \\
                          & RoBERTa-ONE & $\pm$0.0095 & $\pm$0.0087                & $\pm$0.0364 & $\pm$0.0358        & $\pm$0.0382 & $\pm$0.0432         & $\pm$0.0169 & $\pm$0.0162            & $\pm$0.0197 & $\pm$0.0187              \\
                          & Adapter-ONE & $\pm$0.0292 & $\pm$0.0223                & $\pm$0.0207 & $\pm$0.0222        & $\pm$0.0076 & $\pm$0.0063         & $\pm$0.0141 & $\pm$0.0157            & $\pm$0.0074 & $\pm$0.0054              \\
                          & Prompt-ONE  & $\pm$0.0427 & $\pm$0.0991                & $\pm$0.0297 & $\pm$0.0254        & $\pm$0.0386 & $\pm$0.0325         & $\pm$0.0115 & $\pm$0.0100            & $\pm$0.0151 & $\pm$0.0292              \\
                          & DEMIX       & $\pm$0.0329 & $\pm$0.0293                & $\pm$0.0259 & $\pm$0.0283        & $\pm$0.0297 & $\pm$0.0367         & $\pm$0.0336 & $\pm$0.0309            & $\pm$0.0152 & $\pm$0.0165              \\ 
\hline
\multirow{8}{*}{CL}       & RoBERTa-NCL         & $\pm$0.0374 & $\pm$0.0238                & $\pm$0.0156 & $\pm$0.0158        & $\pm$0.0293 & $\pm$0.0349         & $\pm$0.0218 & $\pm$0.0154            & $\pm$0.0130 & $\pm$0.0160              \\
                          & Adapter-NCL & $\pm$0.0250 & $\pm$0.0194                & $\pm$0.0232 & $\pm$0.0184        & $\pm$0.0183 & $\pm$0.0264         & $\pm$0.0136 & $\pm$0.0151            & $\pm$0.0095 & $\pm$0.0137              \\
                          & HAT         & $\pm$0.0264 & $\pm$0.012                 & $\pm$0.0236 & $\pm$0.0251        & $\pm$0.0294 & $\pm$0.0287         & $\pm$0.0106 & $\pm$0.009             & $\pm$0.0078 & $\pm$0.0112              \\
                          %& CLASSIC     & $\pm$0.0375 & $\pm$0.0416                & $\pm$0.0183 & $\pm$0.0146        & $\pm$0.0108 & $\pm$0.0141         & $\pm$0.0276 & $\pm$0.0222            & $\pm$0.0124 & $\pm$0.0108              \\
                          & BCL         & $\pm$0.0255 & $\pm$0.0124                & $\pm$0.0121 & $\pm$0.0105        & $\pm$0.0182 & $\pm$0.0126         & $\pm$0.0100 & $\pm$0.0069            & $\pm$0.0094 & $\pm$0.0032              \\
                          & KD          & $\pm$0.0642 & $\pm$0.0435                & $\pm$0.0295 & $\pm$0.0233        & $\pm$0.0271 & $\pm$0.0267         & $\pm$0.0160 & $\pm$0.0133            & $\pm$0.0117 & $\pm$0.0109              \\
                          & EWC         & $\pm$0.0324 & $\pm$0.0259                & $\pm$0.0281 & $\pm$0.0189        & $\pm$0.0177 & $\pm$0.0196         & $\pm$0.0041 & $\pm$0.0096            & $\pm$0.0079 & $\pm$0.0062              \\
                          & DER++       & $\pm$0.0250 & $\pm$0.0183                & $\pm$0.0231 & $\pm$0.0319        & $\pm$0.0116 & $\pm$0.0163         & $\pm$0.0196 & $\pm$0.0178            & $\pm$0.0126 & $\pm$0.0128              \\
                          & CPT         & $\pm$0.0264 & $\pm$0.0120                & $\pm$0.0236 & $\pm$0.0251        & $\pm$0.0294 & $\pm$0.0287         & $\pm$0.0106 & $\pm$0.0090             & $\pm$0.0078 & $\pm$0.0112              \\
\specialrule{.1em}{.05em}{.05em}
\end{tabular}
}
\caption{Standard deviations of the corresponding metrics of the proposed CPT system and the baselines.}
\label{tab:std}
\end{table*}

\begin{table}[]
%\small
\centering
\resizebox{0.7\columnwidth}{!}{
\begin{tabular}{c||ccc}
\specialrule{.2em}{.1em}{.1em}
\multirow{2}{*}{Model} & \multicolumn{2}{c}{Final Performance}  \\
 & MF1 & Acc \\
\specialrule{.1em}{.05em}{.05em}
CPT (Sequential Adapter)  & $\pm${0.0347}  & $\pm${0.0350}  \\
\hline
CPT (w/o butterfly)  &  $\pm${0.0102}  & $\pm${0.0079}   \\
CPT (w/o masking)  &  $\pm${0.0095}  & $\pm${0.0137}   \\
\hline
CPT  & $\pm${0.0078} &  $\pm${0.0112} \\
\specialrule{.1em}{.05em}{.05em}
%\vspace{-4mm}
\end{tabular}
}
\caption{Standard deviations of the corresponding metrics of the proposed CPT system and the ablations. }
\label{tab:ablation_std}
\end{table}

%\zixuan{TODO: can you specifiy the order? also in the 3.2 main paper, abalation (4) }

\begin{table*}[]
\centering
\resizebox{\textwidth}{!}{%
\begin{tabular}{cc!{\vrule width \lightrulewidth}cccc|cccc|cccc|cccc|cccc} 
\specialrule{.2em}{.1em}{.1em}
        \multirow{3}{*}{Category} & Order & \multicolumn{4}{c|}{\texttt{AI}$\to$\texttt{ACL}$\to$\texttt{Restaurant}$\to$\texttt{AGNews}}                                   & \multicolumn{4}{c|}{\texttt{Restaurant}$\to$\texttt{AI}$\to$\texttt{AGNews}$\to$\texttt{ACL}}                                   & \multicolumn{4}{c|}{\texttt{AI}$\to$\texttt{ACL}$\to$\texttt{AGNews}$\to$\texttt{Restaurant}}                                   & \multicolumn{4}{c|}{\texttt{AGNews}$\to$\texttt{ACL}$\to$\texttt{Restaurant}$\to$\texttt{AI}} & \multicolumn{4}{c}{\textbf{Average}}                                    \\
         & Metric & \multicolumn{2}{c}{Performance} & \multicolumn{2}{c|}{Forget R.} & \multicolumn{2}{c}{Performance} & \multicolumn{2}{c|}{Forget R.} & \multicolumn{2}{c}{Performance} & \multicolumn{2}{c|}{Forget R.} & \multicolumn{2}{c}{Performance} & \multicolumn{2}{c|}{Forget R.}   & \multicolumn{2}{c}{\textbf{Performance}} & \multicolumn{2}{c}{\textbf{Forget R.}} \\
         & Model & MF1 & Acc                   & MF1 & Acc                     & MF1 & Acc                   & MF1 & Acc                     & MF1 & Acc                   & MF1 & Acc                     & MF1 & Acc                   & MF1 & Acc        
                 & MF1 & Acc                   & MF1 & Acc\\ 
\specialrule{.1em}{.05em}{.05em}
\multirow{6}{*}{Non-CL}   &  RoBERTa  & 43.72 & 50.94 & \multicolumn{2}{c|}{---} & 43.72 & 50.94 & \multicolumn{2}{c|}{---} &43.72 & 50.94 & \multicolumn{2}{c|}{---} & 43.72 & 50.94 & \multicolumn{2}{c|}{---} & 43.72 & 50.94 & \multicolumn{2}{c}{---}\\
&  Adapter  & 39.65 & 46.48 & \multicolumn{2}{c|}{---} & 39.65 & 46.48 & \multicolumn{2}{c|}{---} & 39.65 & 46.48 & \multicolumn{2}{c|}{---} & 39.65 & 46.48 & \multicolumn{2}{c|}{---} & 39.65 & 46.48 & \multicolumn{2}{c}{---}\\
&  RoBERTa-ONE  & 44.78 &51.92 & \multicolumn{2}{c|}{---} & 44.78 &51.92 & \multicolumn{2}{c|}{---} &44.78 &51.92 & \multicolumn{2}{c|}{---} &44.78 &51.92 & \multicolumn{2}{c|}{---}&44.78 &51.92 & \multicolumn{2}{c}{---}\\
&  Adapter-ONE  & 45.31 & 51.68 & \multicolumn{2}{c|}{---} & 45.31 & 51.68 & \multicolumn{2}{c|}{---} &45.31 & 51.68 & \multicolumn{2}{c|}{---} &45.31 & 51.68 & \multicolumn{2}{c|}{---}&45.31 & 51.68 & \multicolumn{2}{c}{---}\\
&  Prompt-ONE  & 34.68 & 43.42 &
 \multicolumn{2}{c|}{---} & 34.68 & 43.42 &
 \multicolumn{2}{c|}{---} &34.68 & 43.42 &
 \multicolumn{2}{c|}{---} &34.68 & 43.42 &
 \multicolumn{2}{c|}{---}&34.68 & 43.42 &
 \multicolumn{2}{c}{---}\\
&  DEMIX  & 45.41 & 51.57 &
 \multicolumn{2}{c|}{---} & 45.41 & 51.57 &
 \multicolumn{2}{c|}{---} &45.41 & 51.57 &
 \multicolumn{2}{c|}{---} &45.41 & 51.57 &
 \multicolumn{2}{c|}{---}&45.41 & 51.57 &
 \multicolumn{2}{c}{---}\\
\hline
\multirow{8}{*}{CL}   &  RoBERTa-NCL  & 42.62 & 49.95 & 2.45 & 1.79 & 42.22 & 49.52 & 3.10 & 2.33 &42.88 & 50.11 & 0.29& 0.18 &44.33 & 51.51 &1.76 & 1.21 & 43.01 & 50.28 & 1.90 & 1.38\\
& Adapter-NCL & 44.71 & 51.67 & 1.71 & 1.08 & 44.61 & 51.07 & 1.14 & 1.23 & 44.91 & 51.57 & 1.41 &1.23 & 45.52 & \textbf{52.15} & 0.72& 0.44 & 44.94 & 51.62 & 1.25 & 0.99  \\
& HAT &  45.10 & 51.50 & 1.66 & 1.19 & 43.29 & 49.96 & 2.76 & 2.09 & 46.06 & \textbf{52.07} & 0.50 & 0.21 & 44.94 & 51.45 & 0.86 & 0.25 & 44.85 & 51.25 & 1.45 & 0.93 \\
%CLASSIC  & 42.48 & 49.69 & 1.86 & 1.78 & \textbf{45.73} & 52.19 & {-0.39} & {-0.56} & 45.09 & 51.75 & {-1.19} & {-0.88} & 44.88 & 51.33 & {-0.50} & {-0.09} \\
& BCL  & 43.97 & 50.74 & 2.20 & 1.50 & 45.30 & 51.54 & 0.36 & -0.14 & 45.28 & 51.79 & 0.36 & 0.11 & 45.59 & 51.61 & 0.08 & 0.11 & 45.04 & 51.42 & 0.75 & 0.40\\
& KD & 42.09 & 50.22 & 0.57 & 0.08 &45.18 & \textbf{52.68} &1.22 &0.57 & 42.63 & 50.45 & \textbf{-0.31} & \textbf{-0.56} &42.93 & 50.70 & 1.10& 0.32 & 43.21 & 51.01 & 0.64 & 0.10  \\
& EWC & 43.97 & 50.74 &0.16 & 0.03 & 43.65 & 50.29 & \textbf{-0.29}& \textbf{-0.20} & 45.52 & 51.36 & 0.17& 0.15 & 43.42 & 49.85 & 0.12& 0.10 & 44.14 & 50.56 & 0.04 & 0.02 \\
& DER++  & 44.56 & 50.13 & 2.95 & 2.31 & 44.02 & 49.99 & 1.24 & 1.12 & 43.98 & 50.23 & 1.44 & 1.27 & 44.32 & 50.13 & 1.32 & 1.09 & 44.22 & 50.12 & 1.74 & 1.45 \\
& CPT    & \textbf{46.49} & \textbf{52.47} & \textbf{0.00} & \textbf{0.00} & \textbf{45.71} & 51.71 & 0.00 & 0.00 & \textbf{46.15} & 51.93 & 0.00 & 0.00 & \textbf{45.89} & 51.86 & \textbf{0.00} & \textbf{0.00} & \textbf{46.06} & \textbf{51.99} & \textbf{0.00} & \textbf{0.00} \\
\specialrule{.1em}{.05em}{.05em}
\end{tabular}
}
\caption{CPT performance averaged over all domains after the final post-trained with different orders (averaged over 5 random seeds) and the average of these orders. %\bing{please also highlight the best forgetting rates. Can you also provide columns for averages. }
}
\label{tab:order}

\end{table*}

\section{Illustration of Task Masks}
\label{sec.illustration}
Figure \ref{plugin} illustrates the CPT architecture and the task mask learning. Note that fine-tuning is for evaluating the domain post-training and should not affect any parameters of post-training. 
% Therefore, the end-user {is supposed to} \textbf{download} the model before performing fine-tuning.
During \textbf{continual post-training} (Figure~\ref{plugin} (A)), after training domain/task 1, we obtain its useful neurons indicated by the 1 entries. Before training domain/task 2, those useful neurons for domain 1 are first masked (those previous 1's entries are turned to 0's). After training domain 2, two neurons with 1 are used by the domain. When domain $t$ arrives, all used neurons by domains 1 and 2 are masked before training, i.e., their entries are set to 0. After training domain $t$, we see that domains $t$ and 1 have a shared neuron (the cell with two colors, red and green), which is used by both of domains. After continual post-training, we evaluate CPT by \textbf{individual fine-tuning}. During fine-tuning (Figure~\ref{plugin} (B)), we only make use of those neurons that are useful for domain/task id $t$ (red cells) and freeze all other neurons (grey cells).

\section{Dataset Statistics}
\label{sec:data_stat}

Table \ref{tab:dataset} shows the statistics of the \textit{unlabeled domain datasets} and \textit{end-task classification datasets}.
% \bing{The part can be moved to Appendix}
Note that the full AGNews is very large. We use only its author provided training split as our domain-specific datasets as our \textit{unlabeled AGNews} dataset for continual post-training. The remaining testing set is used as the labeled end-task (\textit{AGNews-FT}). The other three corresponding end task datasets are \textit{SemEval-res} \cite{DBLP:conf/naacl/XuLSY19}, \textit{ACL-ARC} \cite{DBLP:journals/tacl/JurgensKHMJ18}, and \textit{SCIERC}\cite{DBLP:conf/emnlp/LuanHOH18}. % \bing{I do not see the test set size for each fine-tuning end task in the table?} 

\section{Details of the CL baselines}
\label{ap:baselines}
% \zixuan{need update}

\textbf{Non-Continual Learning Baselines}: Each of these baselines builds a separate model for each
task independently.
% , which we call a \textbf{ONE} variant. 
It thus has no CF. 
% \zixuan{may want to emphasis more on our advantage: "unlike many existing CL methods, our method needs no replay memory and no network expansion. Domain identification information is also not needed for end-task learning"}

(1,2) \textbf{RoBERTa, Adapter}~\cite{DBLP:journals/corr/abs-1907-11692,Houlsby2019Parameter} use the original RoBERTa/Adapter for the end-task fine-tuning without any post-training. These are the only two without any post-training. All the following baselines use the masked language model loss (MLM) for post-training. 

(3) \textbf{RoBERTa-ONE} is the existing post-training method in~\cite{DBLP:conf/acl/GururanganMSLBD20}. To our knowledge, the existing post-training systems are all based on the MLM loss. 

(4) \textbf{ Adapter-ONE}~\cite{madotto2020continual,Houlsby2019Parameter} adds small adapter layers between layers of Transformer for post-training. We follow the adapter design in~\cite{madotto2020continual,Houlsby2019Parameter}. An adapter is simply two fully connected layers. During post-training, the Transformer is fixed, only the added adapters are trainable. The bottleneck size (adapter size) is set to 128. 
% following~\cite{Houlsby2019Parameter}. % Note that we only fix the RoBERTa in DAPT. 
During end-task fine-tuning, both RoBERTa and the adapters are trainable to ensure fair comparison.
% Note this is the method in \cite{madotto2020continual}, which is called ``Adapter-NCL''.

(5) \textbf{Prompt-ONE}~\cite{DBLP:conf/emnlp/LesterAC21} adds a sequence of real vector tokens (called virtual tokens or prompt tokens) to the end of the original input sequence. In post-training, RoBERTa (the LM) is fixed and only the prompt tokens are trained. In end-task fine-tuning, both LM and the trained prompt are trainable. We initialize 100 tokens and set the learning rate of the prompt token to 0.3 in post-training, following the setting in \cite{DBLP:conf/emnlp/LesterAC21}. 

(6) \textbf{DEMIX}~\cite{gururangan2021demix} is a recent model to adapt a pre-trained LM with new domains. It adds a new adapter once a new domain arrives (network expansion is needed) and initializes the new adapter with the parameters of the previous trained adapter nearest to the new domain data. They use the perplexity on held-out samples to choose the most probable adapter. 
% For fair comparison, we use the same size as $\{\bm{x}^{\text{sub}}_m\}$ as the held-out samples.

\textbf{Continual Learning (CL) Baselines.}

(7) \textbf{RoBERTa-NCL (Naive continual learning)} is a naive extension of \cite{DBLP:conf/acl/GururanganMSLBD20}, which continually/incrementally post-trains the LM to learn all domains using the MLM loss with no mechanism to deal with forgetting or CF.

(8) \textbf{Adapter-NCL}~\cite{Houlsby2019Parameter} is similar to the Adapter based system. The only difference is that the same set of adapters is shared across all domains, rather than using a new adapter for each new domain.

(9) \textbf{Hard attention to overcome forgetting (HAT)} % ~\cite{ke2021adapting} 
is derived from HAT~\cite{Serra2018overcoming}, the state-of-the-art parameter-isolation based method with almost no forgetting. {\color{black} However, HAT suffers from forgetting in continual post-training due to the catastrophic butterfly effect.}
% However, HAT requires task id information in end-task fine-tuning.\bing{we also need, right? Why highlight this point?} HAT also needs to train an addition task embedding to mask each layer of the network which makes the post-training inefficient \bing{why highlighting this?}. % Since HAT is not directly applicable to Transformer, \citet{ke2021adapting} adapts HAT to BERT by applying it to the adapter layer (thus HAT). We replace the backbone network from BERT with RoBERTa for fair comparison. 

(10) \textbf{BCL}~\cite{ke2021adapting} is a continual learning model that can avoid forgetting and encourage knowledge transfer. It is similar to Adapter-NCL. The difference is that its adapters consist of two modules, one is a capsule network (a new capsule is added once a new domain arrives) to encourage transfer, and the other is similar to HAT to avoid forgetting. Similar to HAT, task/domain information is needed in end-task fine-tuning. We replace the backbone network from BERT with RoBERTa for fair comparison.

(11) \textbf{Knowledge distillation (KD)}~\cite{hinton2015distilling}
minimizes the representational deviation between the learned representation and the new representation in post-training. We
compute the KL divergence between the representations (the output before the masked language model prediction head) of each token of the previous post-trained LM and current LM as the distillation loss.

(12) \textbf{EWC}~\cite{buzzega2020dark} is a popular
regularization-based continual learning method that adopts elastic weights consolidation to add $L_2$ regularization to penalize parameter changes.

(13) \textbf{DER++}~\cite{buzzega2020dark} is a recent replay method using distillation to regularize the new task training. We store 16.4K tokens for each learned domain as the memory, which is the largest memory we can use for the system to run.

\section{Results for Different Domain Orders}
\label{ap:order}

% \zixuan{need update}
Table~\ref{tab:overall_results} in the main paper reported the results for the order \texttt{Restaurant} $\to$ \texttt{AI} $\to$ \texttt{ACL} $\to$ \texttt{AGnews}. We now look at how the order affects the results.
% Due to the computation intensive nature of post-training, we only report the best baseline (NCL) and CPT results with different domain orders. 
Table~\ref{tab:order} shows baselines and CPT's results of 4 different orders. Note that the results for the Non-CL baselines are the same across different orders (and the same as those in Table~\ref{tab:overall_results}) because they are not effected by orders. We can see CPT is always better than other baselines, and achieve 0 forgetting rate, demonstrating the effectiveness of CPT. We also note that some baselines in some sequence has negative forgetting rate, indicating they have some backward transfer (new domain learning helps learned domains). However, their final results are much worse than CPT's.

\section{Standard Deviations}
\label{ap:std}
% \zixuan{need update}

Table~\ref{tab:std} reports the standard deviations of the corresponding results in Table~\ref{tab:overall_results} (in the main paper) of CPT and the considered baselines over 5 runs with random seeds. We can see the results of CPT are stable. Some baselines (e.g., RoBERTa, RoBERTa-ONE) can have quite large standard deviations. 

Table~\ref{tab:ablation_std} reports the standard deviations of the corresponding results in Table~\ref{tab:ablation_results} (in the main paper) of CPT and the considered baselines over 5 runs with random seeds. We can see the results of sequential adapters has a high variance while CPT and other variants are stable. 
% Some baselines (e.g., CPT (random mask) and CPT (-contrast) in Camera) can have quite large standard deviations. %  \bing{no std for results in Table 3?}\zixuan{added}

% \bing{need to mention the other baselines in the table and also explain why some are not used. Also add the those non-CL baselines are not used.}

% \section{Example Appendix}
% \label{sec:appendix}

% This is an appendix.

\end{document}